\pdfoutput=1

\documentclass[11pt]{article}

\usepackage[final]{acl}

\usepackage{times}
\usepackage{latexsym}

\usepackage{multirow}
\usepackage{amsmath}
\usepackage{amssymb}
\usepackage{mathtools}
\usepackage{amsthm}
\usepackage{bm}
\newcommand{\methodname}{{\sc MCmark}}
\usepackage{algorithm}
\usepackage[noend]{algpseudocode}
\usepackage{bbding}

\newtheorem{theorem}{Theorem}[section]

\newtheorem{definition}[theorem]{Definition}

\usepackage{booktabs}

\newcommand{\x}{\bm{x}}

\newcommand{\cK}{\mathcal{K}}

\newcommand{\bbE}{\mathbb{E}}

\usepackage[T1]{fontenc}

\usepackage[utf8]{inputenc}

\usepackage{microtype}

\usepackage{inconsolata}

\usepackage{graphicx}
\usepackage{subcaption}

\title{Improved Unbiased Watermark for Large Language Models}

\author{Ruibo Chen$^*$, Yihan Wu$^*$, Junfeng Guo, Heng Huang \\
  University of Maryland, College Park \\
  \texttt{\{rbchen,ywu42,gjf2023,heng\}@umd.edu} \\}

\begin{document}
\maketitle
\renewcommand{\thefootnote}{\fnsymbol{footnote}}
\footnotetext[1]{Equal contribution.}
\renewcommand{\thefootnote}{\arabic{footnote}}
\begin{abstract}
As artificial intelligence surpasses human capabilities in text generation, the necessity to authenticate the origins of AI-generated content has become paramount. Unbiased watermarks offer a powerful solution by embedding statistical signals into language model-generated text without distorting the quality. In this paper, we introduce \methodname, a family of unbiased, \textbf{M}ulti-\textbf{C}hannel-based watermarks. \methodname\ works by partitioning the model's vocabulary into segments and promoting token probabilities within a selected segment based on a watermark key. We demonstrate that \methodname\ not only preserves the original distribution of the language model but also offers significant improvements in detectability and robustness over existing unbiased watermarks.
Our experiments with widely-used language models demonstrate an improvement in detectability of over 10\% using \methodname, compared to existing state-of-the-art unbiased watermarks.
This advancement underscores \methodname's potential in enhancing the practical application of watermarking in AI-generated texts. Our code is available at \url{https://github.com/RayRuiboChen/MCMark}.
\end{abstract}

\section{Introduction}
As artificial intelligence outstrips human ability in text generation, verifying the authenticity and source of AI-created texts is increasingly crucial. Watermarking of language models \citep{Aaronson2022,kirchenbauer2023watermark,christ2023undetectable,kuditipudi2023robust,hu2023unbiased,wu2023dipmark,chen2024mark,chen2024enhancing,guo2024zeromark} offers an effective method to differentiate between texts generated by humans and machines. This approach involves covertly embedding a statistical signal within the text via a watermark generator that uses specific watermark keys. Detection of this statistical signal is carried out using statistical hypothesis testing, enabling the confirmation of the text's provenance.

Watermarks that do not introduce distortion \citep{Aaronson2022,christ2023undetectable,kuditipudi2023robust,hu2023unbiased,wu2023dipmark,wu2024distortion} are particularly crucial in the watermarking of language models. These watermarks are essential as they are provably capable of maintaining the original output distribution of the language model. The expected distribution of the watermarked model, conditioned on the watermark keys, aligns perfectly with that of the original language model, thus maintaining utility and relevance in practical applications.

However, current unbiased watermarking approaches face practical challenges. For instance, the unbiased watermark \cite{hu2023unbiased} requires access to language model (LM) prompts and APIs, EXP-edit \cite{kuditipudi2023robust} incurs a substantial time cost for detection, and DiPmark \cite{wu2023dipmark} exhibits lower detection accuracy compared to biased watermarks like those reported in \cite{kirchenbauer2023watermark}. Consequently, enhancing the practicality of watermarks is an imperative issue. In our work, we introduce a novel family of unbiased watermarks, termed \methodname, which exhibit enhanced detectability and robustness. In \methodname, we partition the vocabulary into $l$ segments. During the watermark generation process, a watermark key is used to randomly select a segment. Then, the token probabilities within the selected segment are promoted using our \methodname-based unbiased algorithm. During detection, the presence of the watermark is detected by verifying whether the current token corresponds to the segment associated with the watermark key.

Our contribution can be summarized as follows:
\begin{itemize}
    \item We introduced \methodname, a family of unbiased watermarks that provably preserve the output distribution of language models. \methodname\ is adaptable, robust to text modification, and does not require access to prompt and language model APIs during detection.
    
    \item We theoretically demonstrate that when the number of segments equals two, \methodname\ offers superior detectability compared to DiPmark\cite{wu2023dipmark} and STA-1~\cite{mao2024watermark}. We further discuss the trade-offs between detectability and robustness in \methodname.
    
    \item Through comprehensive experiments, we validate the unbiasedness, detectability, and robustness of \methodname\ on popular language models, such as LLAMA-3. Our results show an over 10\% improvement in detectability compared to the state-of-the-art unbiased watermarks.
\end{itemize}

\section{Related Work}
\textbf{Statistical watermarks.} \citet{kirchenbauer2023watermark} refined the statistical watermarking framework initially introduced by \citet{Aaronson2022}, showcasing the efficacy of this technique via comprehensive experiments on large language models. They divided the language model tokens into red and green lists and favored the green list tokens by adjusting their logits with a fixed increment $\delta$. \citet{zhao2023provable} introduced a unigram watermark approach that employs single-gram hashing to generate watermark keys, enhancing the robustness of statistical watermarks. \citet{liu2023semantic} further increased the robustness of statistical watermarking by using the semantics of generated texts as watermark keys. Additionally, \citet{liu2023unforgeable} developed a scheme for unforgeable watermarks that utilizes neural networks to alter token distributions, moving away from conventional watermark keys. Nevertheless, such methods can substantially alter the text distribution, potentially diminishing the quality of the content.

\noindent\textbf{Unbiased watermarks.} To maintain the original output distribution in watermarked content, several researchers have investigated novel approaches for token distribution modification. \citet{Aaronson2022} pioneered an unbiased watermarking method using Gumbel-max to adjust token distribution and employing prefix n-grams as watermark keys. \citet{christ2023undetectable} used inverse sampling for modifying the token distributions of watermarked content on a binary language model with watermark keys based on token positioning. ITS-edit and EXP-edit \cite{kuditipudi2023robust} utilized inverse-sampling and Gumbel-max respectively for modifying the token distributions of watermarked content, with a predetermined watermark key list. \citet{hu2023unbiased} combined inverse-sampling and $\gamma$-reweight strategies for watermarking, though their detection method is not model-agnostic. DiPmark \cite{wu2023dipmark} enhanced the $\gamma$-reweight technique and introduced a model-agnostic detector. STA-1 \cite{mao2024watermark} optimized the quality of the watermarked text under the low-entropy scenarios.

\section{Preliminary}
\paragraph{Notation.} We follow the notations in the previous work \cite{hu2023unbiased,wu2023dipmark,mao2024watermark} to represent the generation task of LLMs.  Let the set of vocabulary tokens be denoted by \( V \) with cardinality \( N = |V| \). We define \( \mathcal{V} \), which includes all possible token sequences of any length including zero. In the context of a language model, token sequences are generated in response to a specific prompt. The probability of generating the next token \( x_{t+1} \) from \( V \), conditioned on the preceding token sequence \( x_1, \ldots, x_t \), is represented as \( P_M(x_{t+1} \mid \x_{1:t}) \).

In the watermark generator, the LLM utilizes a private key \( k\in\cK \) to reweight the distribution from \( P_M(x_{t+1} \mid \x_{1:t}) \) to \( P_{M, w}(x_{t+1} \mid \x_{1:t}, k) \), where \( P_{M, w} \) indicates a watermarked model and the private key \( k \) is randomly selected from a key space \( \cK \) according to a known distribution \( P_\cK(k) \). According to \cite{hu2023unbiased}, an unbiased watermark requires that the expectation of the reweighted distribution equals that of the original distribution, i.e.,
$$\mathbb{E}_{k\sim P_\cK}[P_{M, w}(x_{t+1} \mid \x_{1:t},k)]=P_{M}(x_{t+1} \mid \x_{1:t}).$$

During watermark detection, the user only has access to the watermark key, the reweight strategy, and the generated audio. The detector employs a hypothesis testing approach to ascertain the presence of the watermark signal. The null hypothesis $H_0$ is defined as \textit{``The content is generated without the presence of watermarks"}. The detector adopts a score function based on the watermark key and the reweight strategy, which exhibits statistical bias between the watermarked and unwatermarked token sequences.

\section{Method}
\begin{definition}[Distribution Channel] Given the original LM distribution $P_M(\cdot|x_{1:t})$, the reweighting method and the key space, we define each unique watermarked distribution $P_{M,w}(\cdot|x_{1:t},k)$ as a distribution channel. The set of distribution channels is the set of all possible LM distributions after watermarking, i.e., $\{P_{M,w}(\cdot|x_{1:t},k)|k\in\cK\}$.
\end{definition}
For example, for the inverse-sampling and the Gumbel-max method, the set of distribution channels is $\{\delta(x)|x\in V\}$, where $\delta$ is the Dirac distribution, $\delta(x)$ means the probability of sampling token $x$ is 1. It is easy to see the cardinality of the distribution channel set is $|V|$.

We can define the unbiased watermark from the perspective of distribution channels. Let $\{P_1, \ldots, P_l\}$ be a set of distribution channels. Define $\mathcal{K}_i$ as the subset of watermark keys that satisfy $P_{M,w}(\cdot \mid x_{1:t}, k) = P_i$ for all $k \in \mathcal{K}_i$. An unbiased watermark should meet the following condition: $\forall x \in V$, it should have:
\[
\sum_{i=1}^l P_i(x) \mathbb{E}_{k \sim P_{\mathcal{K}}} \left[\mathbf{1}_{\{k \in \mathcal{K}_i\}}\right] = P_{M}(x\mid x_{1:t}).
\]

\subsection{\methodname\ Overview}
 By utilizing the distribution channels, we can design a new watermarking framework. During generation, the watermark key is used to pseudorandomly select one of these channels as the basis for the next token distribution. For detection, the algorithm verifies whether a given token was generated by our model by assessing whether the token aligns with the recovered distribution channel $i_t$. The detailed algorithms for the generation and detection processes are provided in Alg.~\ref{alg:generation_framework} and Alg.~\ref{alg:detection_framework}.
 
 Based on this framework, we propose \methodname, a family of \textbf{M}ulti-\textbf{C}hannel-based unbiased watermarking algorithms. \methodname\ operates by constructing a set of $l$ distribution channels. We divide the vocabulary $V$ into $l$ equal parts, $V_1, \ldots, V_l$. For each distribution channel $P_i$, we increase the probability of tokens in $V_i$. During detection, given the recovered distribution channel index $i_t$, if the generated token $x_t \in V_{i_t}$, we assume this token is generated by the watermarked distribution. The detailed algorithms for the generation and detection processes are shown in Alg.~\ref{alg:generator} and Alg.~\ref{alg:detector}.

The effectiveness of our proposed watermarking approach hinges on designing optimal distribution channels that maximize $\bbE_{i_t \in \{1,2,\cdots, l\}}\sum_{x \in V_{i_t} }P_{i_t}(x)$, that is if the selected channel is $i_t$,  the probability of generating tokens within $V_{i_t}$ should be maximized to ensure that more tokens can be effectively detected.

In the following sections, we detail the process for obtaining the optimal distribution channels.

\subsection{Finding Unbiased Watermarks as an Optimization Problem}

\begin{figure}
    \centering
    \includegraphics[width=1\linewidth]{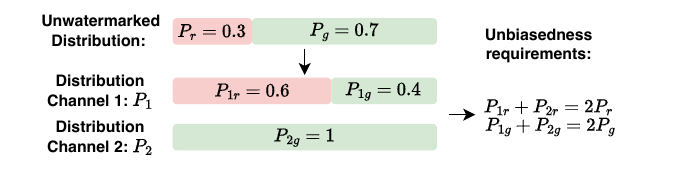}
    \caption{Illustration of the optimal solution for the binary channel example, where we set the red and green list token probabilities to $P_r=0.3$ and $P_g=0.7$ respectively.}
    \label{fig:binary example}
    \vspace{-0.5cm}
\end{figure}

\paragraph{A binary channel example.}
Using the concept of distribution channels, we can explore new unbiased watermarks by identifying sets $\{P_1, \ldots, P_l\}$ and $\{\mathcal{K}_1, \ldots, \mathcal{K}_l\}$. Consider the simplest case where $l=2$ and $\mathbb{E}_{k \sim P_{\mathcal{K}}}[\mathbf{1}_{k \in \mathcal{K}_1}] = \mathbb{E}_{k \sim P_{\mathcal{K}}}[\mathbf{1}_{k \in \mathcal{K}_2}] = \frac{1}{2}$. In this scenario, the watermark possesses two distribution channels with equal probability. To ensure detectability of the watermark (in the presence of a watermark key), the distributions $P_1$ and $P_2$ must be sufficiently distinct. A practical approach involves dividing all tokens into two lists, red and green. In distribution channel $P_1$, we increase the sum of the probability of tokens in the red list ($P_{1r}:=\sum_{x \in V_r}P_1(x)$), and in distribution channel $P_2$, we increase the probability of tokens in the green list ($P_{2g}$). Let $P_r$ and $P_g$ denote the cumulative probabilities of the red and green tokens in the original language model distribution, respectively. The probabilities $P_{1r}, P_{1g}, P_{2r}, P_{2g}$ must satisfy the following constraints for maintaining unbiased properties:
\begin{equation}\label{eqn:simple constrain}
\left\{\begin{array}{l}
    P_{1r} + P_{1g} = 1, \\
    P_{2r} + P_{2g} = 1, \\
    P_{1r} + P_{2r} = 2P_r, \\
    P_{1g} + P_{2g} = 2P_g.
\end{array}\right.
\end{equation}
The first two constraints ensure that the sum of probabilities within a distribution equals 1, while the last two constraints are necessary to uphold the unbiased nature of the watermark.

In order to maximize the detection efficiency, we expect the variation between $P_{1}$ and $P_{2}$ to be as large as possible. Thus our optimization objective is $$\max P_{1r}+P_{2g}.$$
Assuming w.l.o.g. $P_{r}\leq 0.5$, it is easy to identify the optimal solution is $P_{1r}=2P_{r}$ and $P_{2g}=1$ with the constrains in Eq.~\ref{eqn:simple constrain}.
See Figure~\ref{fig:binary example} for an illustration of the optimal solution. In this configuration, the probabilities of all red tokens are doubled in the first distribution channel, while the probabilities of all green tokens are doubled in both channels.

\noindent\textbf{Optimization problem.}
We can generalize the binary case ($l=2$) to more complex scenarios. Consider $\mathbb{E}_{k \sim P_{\mathcal{K}}}[\mathbf{1}_{k \in \mathcal{K}_1}] = \cdots = \mathbb{E}_{k \sim P_{\mathcal{K}}}[\mathbf{1}_{k \in \mathcal{K}_l}] = \frac{1}{l}$. Similarly to the $l=2$ case, we can divide the token set into $l$ equal parts, $V_1, \ldots, V_l$, where $|V_i| = \frac{|V|}{l}$. For each distribution channel $P_i$, we increase the probability of tokens in $V_i$. Denoting by $P_{i,V_j}:=\sum_{x \in V_j}P_i(x)$ the probability of part $V_j$ in channel $P_i$, and $P_{V_j}$ the sum of token probabilities within $V_j$ (i.e. $\mathbb{E}_{x \sim P_M}[\mathbf{1}_{x \in V_j}]$), we formulate the following optimization problem:

\begin{equation}
\label{eqn:constraint}
\begin{aligned}
&\max \sum_{i=1}^l P_{i,V_i}, \\
&\text{s.t.}
\begin{cases}
    \sum_{j=1}^l P_{i,V_j} = 1, \, &\forall i = 1, \ldots, l, \\
    \sum_{i=1}^l P_{i,V_j} = l P_{V_j}, \, &\forall j = 1, \ldots, l.
\end{cases}
\end{aligned}
\end{equation}

The constraint $\sum_{j=1}^l P_{i,V_j} = 1$ ensures that the total probability within each distribution channel $P_i$ sums to 1. This requirement guarantees that each $P_i$ represents a valid probability distribution. Additionally, the constraint $\sum_{i=1}^l P_{i,V_j} = lP_{V_j}$ ensures that the expected probability of $V_j$ across all distribution channels equals the original probability of $V_j$ in the model distribution $P_M$. This maintains the unbiased nature of the watermark.

\begin{figure}
    \centering
    \includegraphics[width=1\linewidth]{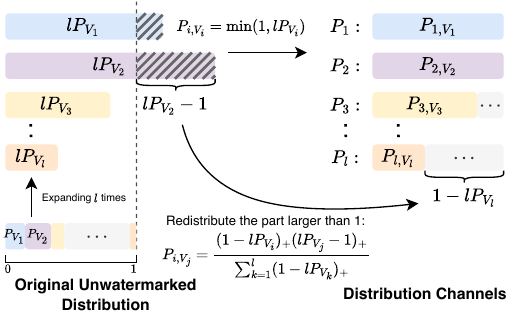}
    \caption{Illustration of the optimal solution (Eq.~\ref{eqn:optimal solution}) for creating the distribution channels for \methodname.}
    \label{fig:optimal solution}
    \vspace{-0.5cm}
\end{figure}

\paragraph{Optimization Solution.}
Given that $\max \sum_{i=1}^l P_{i,V_i} \leq \sum_{i=1}^l \max P_{i,V_i}$, we first calculate each $\max P_{i,V_i}$ individually and then demonstrate the feasibility of $\sum_{i=1}^l \max P_{i,V_i}$. With the constraint $\sum_{i=1}^l P_{i,V_j} = lP_{V_j}$, we have $P_{i,V_i} \leq \min\{1, lP_{V_i}\}$ and $\max P_{i,V_i} = \min\{1, lP_{V_i}\}$. Therefore, we obtain:
\[
\max \sum_{i=1}^l P_{i,V_i} \leq \sum_{i=1}^l \min\{1, lP_{V_i}\}.
\]
We now show that $\sum_{i=1}^l \min\{1, lP_{V_i}\}$ is feasible. We propose one solution:

\small
\begin{equation}
\label{eqn:optimal solution}
P_{i,V_j} =\begin{cases}
      \begin{array}{ll}
            \min\{1, lP_{V_i}\},&\text{if } i = j, \\
            \frac{(1-lP_{V_i})_{+}(lP_{V_j}-1)_{+}}{\sum_{k=1}^{l}(1-lP_{V_k})_{+}},&\text{if } i\neq j.
        \end{array}
\end{cases}
\end{equation} \normalsize

where $(1-lP_{V_i})_{+}:=\max\{0,1-lP_{V_i}\}$ and $(lP_{V_j}-1)_{+}:=\max\{0,lP_{V_j}-1\}$. Please refer Figure~\ref{fig:optimal solution} for an illustration of the optimal solution. The token probabilities within a channel are given by
$P_{i}(x)=\frac{P_{i,V_j}}{P_{V_j}} P_M(x|\bm{x}_{1:t}) ,\forall x \in V_j.$

\subsection{Watermarking Methodology}
With the optimization solution defined, we can establish our watermarking algorithm. Following the approach outlined in \cite{kirchenbauer2023watermark}, we utilize a fixed secret key $\mathsf{sk}$ and the n-gram preceding content $x_{t-n:t-1}$ as the watermark key to generate the token $x_t$. During watermark generation, we first split the vocabulary into $l$ parts and compute the set of distribution channels according to Eq.~\ref{eqn:optimal solution}. We then pseudorandomly select a probability channel based on the watermark key and sample the next token from the selected probability channel. The detailed algorithm is presented in Algorithm~\ref{alg:generator}.

During detection, we are given the generated content and the secret key. With this information, we can recover the index $i_t$ of the probability channel used for generating the token $x_t$ at step $t$. As in $P_{i_t}$, we increase the probability of tokens in $V_{i_t}$, thus we can detect the watermark signal by checking whether $x_t$ is in $V_{i_t}$ or not. We define a hypothesis test with null hypothesis $H_0$: the content is generated without watermarking. Denoted by $\x_{1:T}$ a given content sequence, we can use the test statistic $\Phi(\x_{1:T})=\sum_{t=1}^T\bm{1}_{x_t\in V_{i_t}},$ where $i_t$ is the index of the probability channel at step $t$ recovered from the watermark keys.

Under the null hypothesis, $\Phi(\x_{1:T})$ follows a binomial distribution with a success rate of $\frac{1}{l}$. Thus, we have the following tail bound:

\small\begin{equation*}
\Pr(\Phi(\x_{1:T}) \geq z) = \sum_{i = \lceil z \rceil}^{T} \binom{T}{i} \left(\frac{1}{l}\right)^i \left(\frac{l-1}{l}\right)^{T-i}\end{equation*}
\normalsize

The theoretical false positive rate is therefore:

\begin{equation}
    \sum_{i = \Phi(\x_{1:T})}^{T} \binom{T}{i} \left(\frac{1}{l}\right)^i \left(\frac{l-1}{l}\right)^{T-i}
    \label{eq:fpr}
\end{equation}
\begin{algorithm}[h]
\caption{\methodname\ generator.}\label{alg:generator}
\begin{algorithmic}[1]
\State \textbf{Input:} pretrained LM $P_M$, secret key $\textsf{sk}$, prompt $\bm{x}_{-m:0}$, generate length $T\in\mathbb{N}$.
\For{$t=1,\dots,T$}
\State Split the vocabulary into $l$ parts $\{V_1,...,V_l\}$.
        \State Get the probability distribution of $t$-th token $P_M(\cdot\mid\bm{x}_{-m:t-1})$.
        \State Get $P_{V_j}=\sum\limits_{x \in V_j}P_M(x\mid\bm{x}_{-m:t-1}),\ j=1,...,l$
        \State Generate the distribution channels $\{P_1,...,P_l\}$ by Eq.~\ref{eqn:optimal solution}
        \State Pseudorandomly select a probability channel $P_i$ based on the watermark key $(\textsf{sk},\x_{t-n:t-1})$.
    \State Sample the next token $x_{i}$ from $P_i$.
\EndFor
\State \textbf{return} $\bm{x}_{1:T}$.
\end{algorithmic}
\end{algorithm}

\begin{algorithm}[h]
\caption{\methodname\ detector.}\label{alg:detector}
\begin{algorithmic}[1]
\State \textbf{Input:} pretrained LM $P_M$, generated tokens $\bm{x}_{1:T}$, false positive rate threshold $p_0$.
\State Initialize $\Phi=0$
\For{$t=1,\dots,T$}
        \State Recover the index of the selected distribution channel $i_t$ based on the watermark key.
    \State $\Phi=\Phi+\bm{1}_{x_t\in V_{i_t}}$.
\EndFor
\State Get theoretical false positive rate $p$ by Eq.~\ref{eq:fpr}.
\If{$p\leq p_0$}
\State \textbf{return} $\bm{x}_{1:T}$ is watermarked.
\Else
\State \textbf{return} $\bm{x}_{1:T}$ is not watermarked.
\EndIf
\end{algorithmic}
\end{algorithm}
\begin{figure*}[t]
    \centering
    \includegraphics[width=0.9\linewidth]{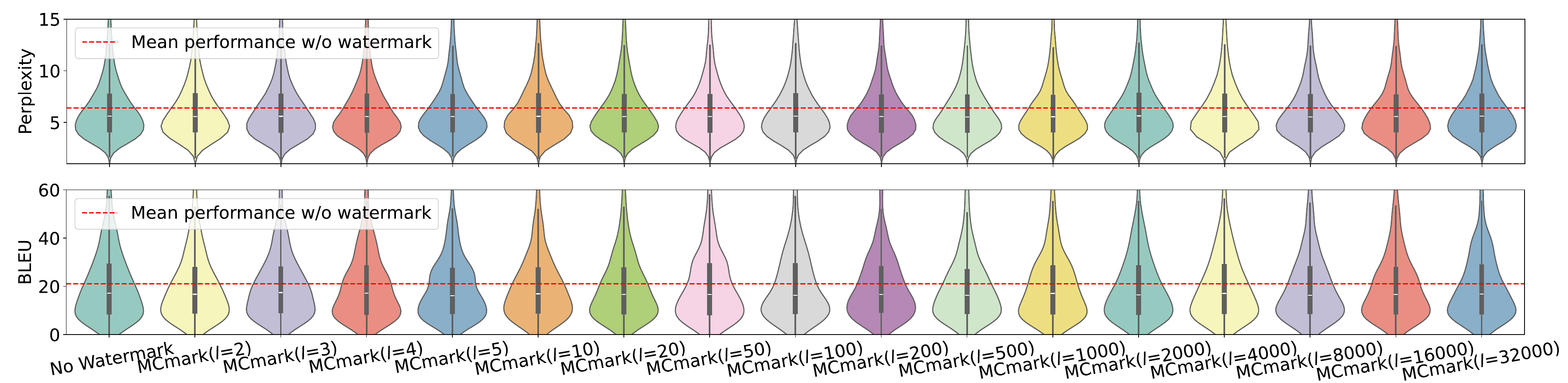}
    \caption{Validating unbiased property of \methodname. \textbf{Top}: text summarization task with perplexity metric. \textbf{bottom}: machine translation task with BLEU metric.}
    \label{fig:Unbiased}
\end{figure*}

\begin{table*}[t]
\centering
\caption{Detectability comparison with LLama2 and C4 dataset on text generation task. Notice, as the detectors of ITS-edit and EXP-edit do not provide a theoretical guarantee of the false positive rate, we report the true positive rate at the empirical false positive rate instead. '-' refers to unavailable results. The lack of results for ITS-edit and EXP-edit at an FPR of 0.01\% is because, in their original setting, the lowest achievable FPR does not reach 0.01\%.}
\label{tab:main_res}
\scalebox{0.9}{
\begin{tabular}{l|cccc}
\toprule
 & TPR@FPR=1\%$\uparrow$ & TPR@FPR=0.1\%$\uparrow$ & TPR@FPR=0.01\%$\uparrow$ & Median p-value$\downarrow$ \\ \midrule
KGW($\delta$=0.5) & 38.78\% & 19.16\% & 10.20\% & 2.809e-2 \\
KGW($\delta$=1.0) & 86.88\% & 74.44\% & 61.55\% & 5.494e-6 \\
KGW($\delta$=1.5) & 96.69\% & 93.83\% & 90.08\% & 1.556e-12 \\
KGW($\delta$=2.0) & 99.34\% & 98.79\% & 97.79\% & 6.580e-22 \\
Unigram($\delta$=0.5) & 78.63\% & 63.51\% & 47.54\% & 1.607e-4 \\
Unigram($\delta$=1.0) & 96.99\% & 92.59\% & 88.08\% & 1.745e-9 \\
Unigram($\delta$=1.5) & 98.94\% & 97.54\% & 96.13\% & 1.051e-16 \\
Unigram($\delta$=2.0) & 99.88\% & 99.52\% & 98.93\% & 5.387e-25 \\ \midrule
ITS-edit & 61.77\% & 54.49\% & - & 4.000e-4 \\
EXP-edit & 89.01\% & 86.35\% & - & 2.000e-4 \\
$\gamma$-reweight & 89.17\% & 81.79\% & 75.83\% & 4.467e-8 \\
DiPmark($\alpha$=0.4) & 87.66\% & 78.77\% & 71.77\% & 1.236e-7 \\
DiPmark($\alpha$=0.3) & 81.88\% & 69.88\% & 61.65\% & 5.284e-6 \\
STA-1 & 84.93\% & 71.58\% & 57.76\% & 2.656e-5 \\
\midrule
\methodname($l$=20) & \textbf{98.96\%} & \textbf{98.38\%} & \textbf{97.69\%} & \textbf{8.098e-30} \\ \bottomrule
\end{tabular}
}
\end{table*}
\subsection{Theoretical analysis}
During the watermark detection process, situations may arise where the current token $x_t$ is sampled from the distribution channel $P_{i_t}$ but does not belong to the designated subset $V_{i_t}$. This discrepancy occurs when $\mathbb{E}_{x \sim P_M}[\mathbf{1}_{x \in V_{i_t}}] < \frac{1}{l}$. Under such conditions, detection of the watermark signal is not possible even though $x_t$ is generated from the watermarked distribution. This leads us to define the true-negative rate, which quantifies the frequency of these undetectable instances, as follows:

\begin{definition}[Expected True-Negative Rate]
We define the true-negative rate, denoted as $P_{TN}(x, P_{M,w}(\cdot \mid k), \Phi)$, as the probability that a token is generated from the watermarked distribution (true) but cannot be detected by the watermark detector (negative): $P_{TN}(x, P_{M,w}(\cdot \mid k), \Phi) := \Pr(\{x \sim P_{M,w}(\cdot \mid k), \Phi(x) = 0\}).$
The expected true-negative rate is then defined as
\[
E_{TN} := \mathbb{E}_{k \sim P_{\mathcal{K}}}[P_{TN}(x, P_{M,w}(\cdot \mid k), \Phi)].
\]
\end{definition}

The true-negative rate is relevant for many statistical watermarks, including Soft watermark~\cite{kirchenbauer2023watermark}, $\gamma$-reweight~\cite{hu2023unbiased}, DiPmark~\cite{wu2023dipmark}, and STA-1~\cite{mao2024watermark}. Notably, $\gamma$-reweight is a special case of DiPmark. In the subsequent analysis, we will compare the expected true-negative rates of DiPmark, STA-1, and our proposed method.

Both DiPmark and STA-1 implement a red-green list strategy. We denote the red list probability as $P_{V_r} := \mathbb{E}_{x \sim P_M}[\mathbf{1}_{x \in V_{r}}]$. The expected true negative rate of DiPmark ($E_{TN}^{\text{DiP}}$), STA-1 ($E_{TN}^{\text{STA}}$) and \methodname\ ($E_{TN}^{\methodname}$) are given by:

\small\begin{equation*}
    \begin{cases}
      \begin{array}{ll}
E_{TN}^{\text{DiP}} = \max\{P_{V_r} - \alpha, 0\} + \max\{P_{V_r} - (1 - \alpha), 0\}.\\
E_{TN}^{\text{STA}} = P_{V_r}^2.\\
E_{TN}^{\methodname} = \sum_{i=1}^l \max\{0, 1/l - P_{V_i}\}.
\end{array}
\end{cases}
\end{equation*}\normalsize

If the probabilities \( P_{V_i} \) are evenly distributed, that is, \( P_{V_i} = \frac{1}{l} \), then the expected true-negative rate for \methodname, \( E_{TN}^{\methodname} \), equals zero, and watermark signals are embedded uniformly across all tokens. In practice, increasing \( l \) tends to result in more unevenly distributed \( P_{V_i} \). In the most extreme case, when \( l = |V| \), each segment \( V_i \) contains exactly one token, which represents the most uneven distribution of \( P_{V_i} \) and thus the expected true-negative rate will be large.

To compare with DiPmark and STA-1, we consider a special case of \methodname\ where $l=2$, meaning there are two distribution channels, and the vocabulary is segmented into a red and a green list. In this scenario, the expected true-negative rate for \methodname\ simplifies to:

\small\begin{equation}
\begin{split}
E_{TN}^{\methodname} &= \max\left\{0, \frac{1}{2} - P_{V_r}\right\}
\\ &+ \max\left\{0, P_{V_r} - \frac{1}{2}\right\}
= \left|\frac{1}{2} - P_{V_r}\right|.
\end{split}
\end{equation}\normalsize

\noindent Note that all of $E_{TN}^{\text{DiP}}$, $E_{TN}^{\text{STA}}$, and $E_{TN}^{\methodname}$ are correlated with $P_{V_r}$. Assuming a uniform distribution of $P_{V_r}$ on $[0,1]$, we compute $\mathbb{E}_{P_{V_r}}[E_{TN}^{\text{DiP}}]$, $\mathbb{E}_{P_{V_r}}[E_{TN}^{\text{STA}}]$, and $\mathbb{E}_{P_{V_r}}[E_{TN}^{\methodname}]$ as $(\alpha - \frac{1}{2})^2 + \frac{1}{4}$, $\frac{1}{3}$, and $\frac{1}{4}$, respectively. This implies that \methodname\ achieves superior detectability compared to DiPmark and STA-1, as indicated by the lower expected true negative rate. Furthermore, the variances of $E_{TN}^{\text{DiP}}$, $E_{TN}^{\text{STA}}$, and $E_{TN}^{\methodname}$ are calculated as $\frac{5}{48} - (\alpha - \frac{1}{2})^2 \left[ (\alpha + \frac{1}{6})^2 + \frac{1}{18} \right]$, $\frac{4}{45}$, and $\frac{1}{48}$, respectively. Given that in DiPmark $\alpha \leq \frac{1}{2}$, \methodname\ also achieves the minimum variance among all three methods. This indicates that \methodname\ can more consistently generate watermarked sentences with a low true negative rate.

\begin{figure*}[t]
    \centering
    \includegraphics[width=1\linewidth]{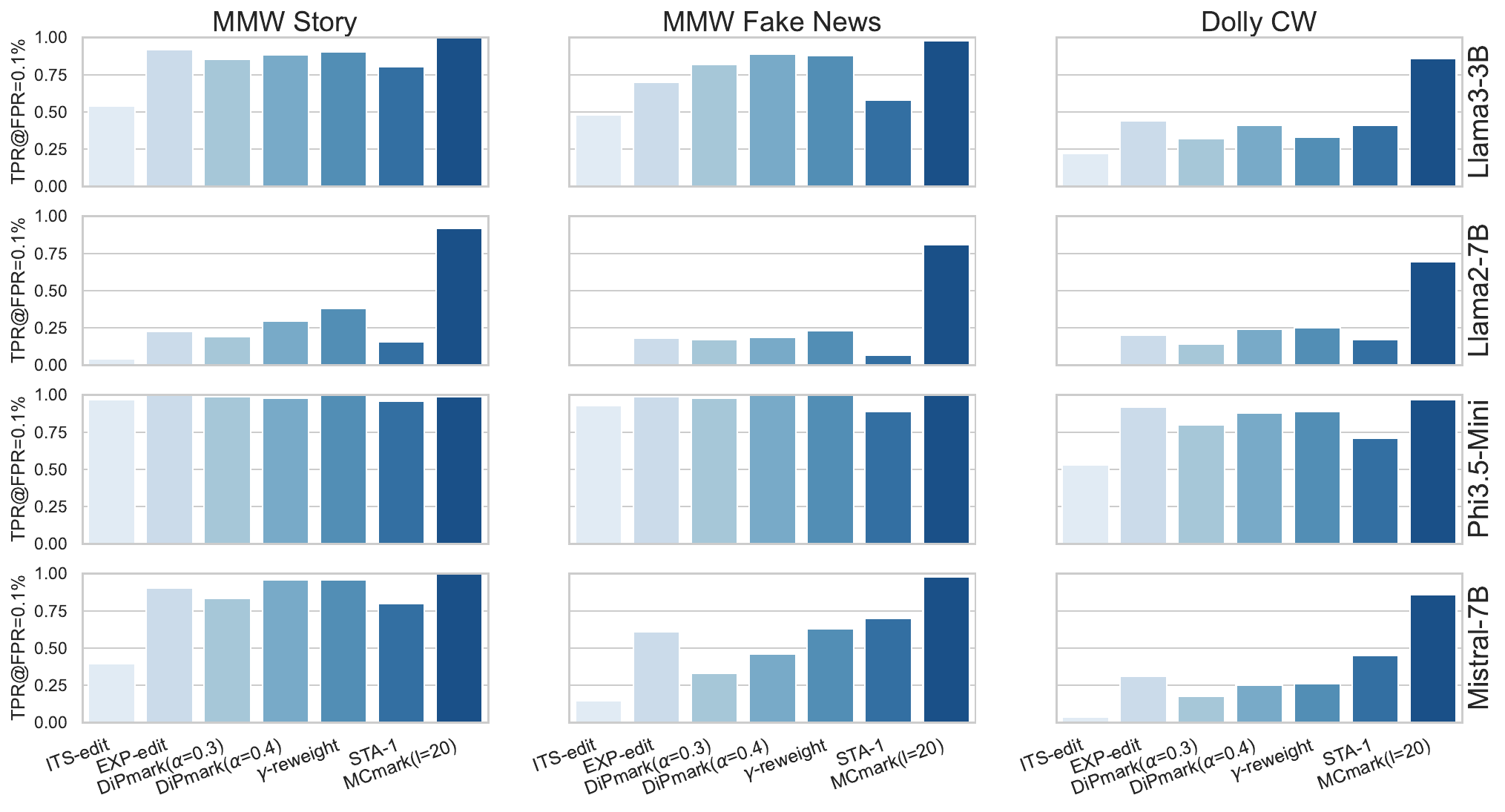}
    \vspace{-0.5cm}
    \caption{Comparative analysis of \methodname\ against SOTA unbiased watermarks across various language models and datasets on watermark detectability.}
    \vspace{-0.5cm}
    \label{fig:detectability comparison}
\end{figure*}

\subsection{Robustness-detectability trade-off}\label{sec:trade-off}

An adversary may attempt to alter the output token to disrupt the watermark detection. In \methodname\ detection, if a token $x_t$ is modified to $x_t'$ and $x_t' \notin V_{i_t}$, the watermark signal is effectively removed. Consequently, the probability that a watermark is removed due to such an alteration is given by $\frac{|V_{i_t}|}{|V|}$.
Given that $|V_{i_t}| = \frac{|V|}{l}$, the probability that a watermark is removed simplifies to $\frac{1}{l}$
Therefore, increasing $l$ decreases the robustness of the watermark, as it increases the likelihood that an adversary can successfully remove the watermark by modifying the token.

On the other hand, we show that moderately increasing \( l \) can enhance the detectability of the watermark (see Appendix~\ref{sec:full version} for a detailed discussion). Thus, we identify a fundamental trade-off: increasing the number of distribution channels $l$ enhances the detectability of the watermark, yet it simultaneously reduces its robustness. We empirically validate our analysis in Figure~\ref{fig:trade-off1} and ~\ref{fig:trade-off2}.

\section{Experiments}

The experiments consist of two main parts. First, we validate that our proposed method is unbiased by demonstrating that its output quality for machine translation and text summarization tasks is similar to the baseline without watermarking. Second, we showcase the effectiveness of \methodname through comprehensive experiments on text generation.
After that, we discuss the robustness of our method and the detectability-robustness trade-off in \methodname. The detailed experimental settings can be found in Appendix~\ref{sec:settings}.

\subsection{Unbiased Property Validation}

Since \methodname\ is \textbf{provably} unbiased, we use this empirical experiment as a support for this property. We follow the evaluation process of \cite{hu2023unbiased}. Specifically for \methodname, we select number of the distribution channels $l$ from the set \{2, 3, 4, 5, 10, 20, 50, 100, 200, 500, 1000, 2000, 4000, 8000, 16000, 32000\}.
Upon examining Figure~\ref{fig:Unbiased}, we find across all $l$ values, the BLEU scores in the machine translation tasks and the perplexity values in the text summarization tasks remain consistently similar between \methodname\ and the original language model. In appendix Table~\ref{tab:unbiased_ts} and Table~\ref{tab:unbiased_mt}, we provide additional unbiasedness evaluation on both biased and unbiased watermarks, the results indicate that \methodname\ can preserve the LM distribution compared to the biased watermarks.

\subsection{Detectability}

Following the evaluation metric of the previous works~\cite{kuditipudi2023robust,wu2023dipmark}, we report the true positive rate at guaranteed false positive rates, i.e., TPR@FPR=$\{1\%, 0.1\%, 0.01\%\}$. Notice, as the detectors of ITS-edit and EXP-edit do not provide a theoretical guarantee, we report the true positive rate at the empirical false positive rate following their original setting. From Table~\ref{tab:main_res}, we see that \methodname\ achieved the best detectability comparing with all unbiased watermarks, at least 14\% improvement on all TPR@FPR metrics. Besides, \methodname\ outperformed the biased watermarking algorithm KGW and Unigram when $\delta \in \{0.5,1.0, 1.5\}$, and achieved comparable performance with them when $\delta=2.0$.
In Figure~\ref{fig:detectability comparison}, we present a comparative analysis of \methodname\ against SOTA unbiased watermarks across various language models and datasets. Our method, represented by the last bar in each plot, consistently outperforms all comparisons across the board. Further experimental results are detailed in Figure~\ref{fig:detectability comparison2}.

\begin{table*}[t]
\centering
\caption{Robustness comparison of unbiased watermarks, we use metrics TPR@FPR=0.1\% and the median p-value.}
\label{tab:robustness}
\begin{tabular}{l|cc|cc|cc}
\toprule
 & \multicolumn{2}{c|}{$\epsilon$=0.05} & \multicolumn{2}{c|}{$\epsilon$=0.1} & \multicolumn{2}{c}{$\epsilon$=0.2} \\
 & TPR$\uparrow$ & p-value$\downarrow$ & TPR$\uparrow$ & p-value$\downarrow$ & TPR$\uparrow$ & p-value$\downarrow$ \\ \midrule
ITS-edit &  50.40\%  & 7.998e-4 & 43.69\% & 6.399e-3 & 33.90\% & 4.839e-2 \\
EXP-edit & 81.35\% & 2.000e-4 & 78.27\% & 2.000e-4 & 74.88\% & 2.000e-4 \\
$\gamma$-reweight & 72.38\% & 4.241e-6 & 59.40\% & 1.975e-4 & 31.07\% & 1.195e-2 \\
DiPmark($\alpha$=0.4) & 69.63\% & 8.154e-6 & 58.13\% & 1.975e-4 & 29.06\% & 1.996e-2 \\
DiPmark($\alpha$=0.3) & 59.53\% & 1.363e-4 & 46.24\% & 2.123e-3 & 20.59\% & 4.059e-2 \\
STA-1 & 60.84\% & 2.253e-4 & 47.15\% & 1.462e-3 & 21.35\% & 1.843e-2 \\
\midrule
\methodname($l$=20) & \textbf{97.11\%} & \textbf{6.068e-23} & \textbf{96.07\%} & \textbf{2.140e-18} & \textbf{88.79\%} & \textbf{3.731e-10} \\ \bottomrule
\end{tabular}
\end{table*}

\begin{table*}[t]
    \centering
    \caption{Robustness comparison of unbiased watermarks under GPT rephrasing attack, we report TPR@FPR=\{5\%,1\%,0.1\%,0.01\%\}, the median p-value, and AUC.}
    \label{tab:gpt_rephrase_detection}
    \begin{tabular}{lcccccc}
        \toprule
        \multirow{2}{*}{GPT rephrasing} &
        \multicolumn{4}{c}{TPR @ FPR (\%)} &
        \multirow{2}{*}{Median \(p\)-value\(\,\downarrow\)} &
        \multirow{2}{*}{AUC\(\,\uparrow\)} \\
        \cmidrule(lr){2-5}
        & 5\% & 1\% & 0.1\% & 0.01\% & & \\
        \midrule
        ITS-edit              &  6.3\% &  2.3\% & 1.0\% &  --   & \(4.721\times10^{-1}\) & 0.5138 \\
        EXP-edit              & 26.4\% & 17.9\% &11.9\% &  --   & \(2.296\times10^{-1}\) & 0.6879 \\
        \(\gamma\)-reweight   & 13.4\% &  6.6\% & 2.2\% & 0.9\% & \(5.635\times10^{-1}\) & 0.5019 \\
        DiPmark(\(\alpha=0.4\)) & 14.9\% &  6.4\% & 2.4\% & 0.7\% & \(6.031\times10^{-1}\) & 0.4921 \\
        DiPmark(\(\alpha=0.3\)) &  9.9\% &  3.7\% & 1.0\% & 0.3\% & \(6.629\times10^{-1}\) & 0.4738 \\
        STA-1                 & 24.0\% & 11.6\% & 4.5\% & 1.8\% & \(2.317\times10^{-1}\) & 0.6850 \\
        \midrule
        \textbf{MCmark(\(l=20\))} & \textbf{62.5\%} & \textbf{48.0\%} & \textbf{31.7\%} & \textbf{20.2\%} & \textbf{\(1.256\times10^{-2}\)} & \textbf{0.8592} \\
        \bottomrule
    \end{tabular}
\end{table*}

\begin{table*}[t]
    \centering
    \caption{
    Robustness comparison of unbiased watermarks under GPT back-translation attack, we report TPR@FPR=\{5\%,1\%,0.1\%,0.01\%\}, the median p-value, and AUC.}
    \label{tab:gpt_back_translation_detection}
    \begin{tabular}{lcccccc}
        \toprule
        \multirow{2}{*}{GPT back translation} &
        \multicolumn{4}{c}{TPR @ FPR (\%)} &
        \multirow{2}{*}{Median \(p\)-value\(\,\downarrow\)} &
        \multirow{2}{*}{AUC\(\,\uparrow\)} \\
        \cmidrule(lr){2-5}
        & 5\% & 1\% & 0.1\% & 0.01\% & & \\
        \midrule
        ITS-edit               &  6.2\% &  2.3\% &  1.0\% &  --   & \(4.710\times10^{-1}\) & 0.5136 \\
        EXP-edit               & 73.9\% & 66.8\% & 60.4\% &  --   & \(2.000\times10^{-4}\) & 0.8841 \\
        \(\gamma\)-reweight    & 64.5\% & 50.8\% & 32.2\% & 18.0\% & \(9.202\times10^{-3}\) & 0.8520 \\
        DiPmark(\(\alpha=0.4\))& 60.4\% & 46.8\% & 30.8\% & 19.8\% & \(1.378\times10^{-2}\) & 0.8352 \\
        DiPmark(\(\alpha=0.3\))& 52.8\% & 35.5\% & 19.8\% &  9.8\% & \(9.829\times10^{-2}\) & 0.7859 \\
        STA-1                  & 66.5\% & 45.1\% & 24.0\% & 11.1\% & \(1.509\times10^{-2}\) & 0.8926 \\
        \midrule
        \textbf{MCmark(\(l=20\))} & \textbf{94.6\%} & \textbf{91.6\%} & \textbf{86.8\%} & \textbf{81.2\%} & \textbf{\(8.572\times10^{-10}\)} & \textbf{0.9796} \\
        \bottomrule
    \end{tabular}
\end{table*}

\begin{table*}[t]
    \centering
    \caption{Robustness comparison of unbiased watermarks under DIPPER attack, we report TPR@FPR=\{5\%,1\%,0.1\%,0.01\%\}, the median p-value, and AUC.}
    \label{tab:dipper_detection}
    \begin{tabular}{lcccccc}
        \toprule
        \multirow{2}{*}{DIPPER} &
        \multicolumn{4}{c}{TPR @ FPR (\%)} &
        \multirow{2}{*}{Median \(p\)-value\(\,\downarrow\)} &
        \multirow{2}{*}{AUC\(\,\uparrow\)} \\
        \cmidrule(lr){2-5}
        & 5\% & 1\% & 0.1\% & 0.01\% & & \\
        \midrule
        ITS-edit               &  5.1\% &  1.3\% & 0.1\% &  --   & \(5.206\times10^{-1}\) & 0.4898 \\
        EXP-edit               & 16.6\% &  9.2\% & 3.5\% &  --   & \(3.735\times10^{-1}\) & 0.5976 \\
        \(\gamma\)-reweight    &  1.8\% &  0.4\% & 0.0\% & 0.0\% & \(9.745\times10^{-1}\) & 0.3124 \\
        DiPmark(\(\alpha=0.4\))&  1.7\% &  0.3\% & 0.0\% & 0.0\% & \(9.634\times10^{-1}\) & 0.3314 \\
        DiPmark(\(\alpha=0.3\))&  1.3\% &  0.6\% & 0.0\% & 0.0\% & \(9.780\times10^{-1}\) & 0.3191 \\
        STA-1                  & 11.6\% &  3.4\% & 0.6\% & 0.1\% & \(3.652\times10^{-1}\) & 0.5917 \\
        \midrule
        \textbf{MCmark(\(l=20\))} & \textbf{24.1\%} & \textbf{11.0\%} & \textbf{3.6\%} & \textbf{1.0\%} & \textbf{\(1.976\times10^{-1}\)} & \textbf{0.6950} \\
        \bottomrule
    \end{tabular}
\end{table*}

\begin{figure}[t]
    \centering
    \includegraphics[width=0.85\linewidth]{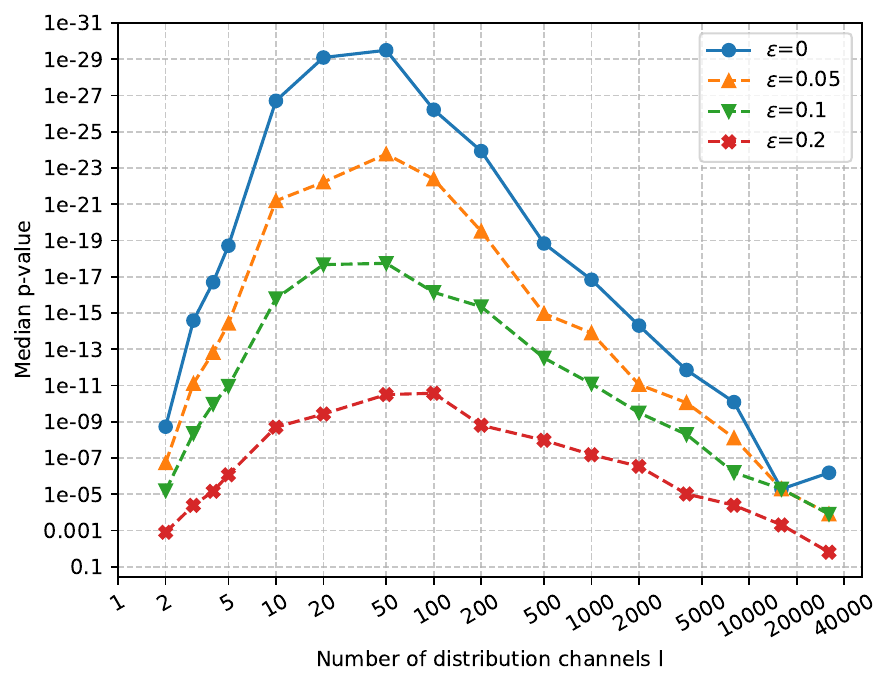}
    \caption{Median p-value vs number of distribution channels $l$ in \methodname.}
    \label{fig:trade-off1}
    \vspace{-0.5cm}
\end{figure}

\begin{figure}[t]
    \centering
    \includegraphics[width=0.85\linewidth]{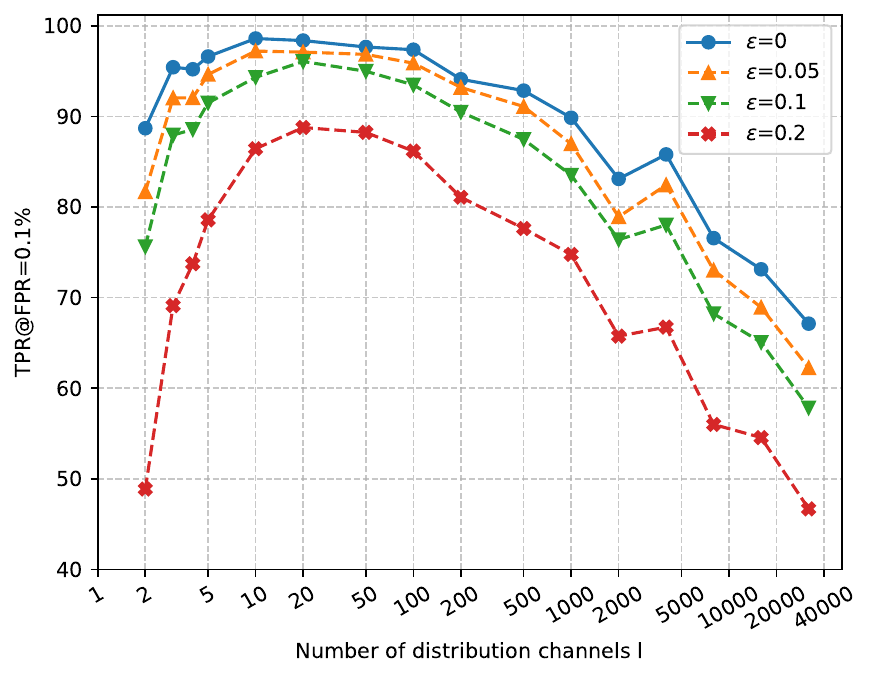}
    \caption{TPR@FPR=0.1\% vs number of distribution channels $l$ in \methodname.}    \label{fig:trade-off2}
     \vspace{-0.5cm}
\end{figure}
\paragraph{Time Efficiency.}
Similar to the KGW watermark, Unigram, and DiPmark, the time cost introduced by the \methodname\ generator occurs only during the modification of token probabilities in the generation process. Additionally, the \methodname\ detector is model-agnostic. Empirically, the time efficiency of the \methodname\ method matches that of KGW, Unigram, $\gamma$-reweight, DiPmark, and STA-1. Notably, the detectors of ITS-edit and EXP-edit require thousands of inferences~\cite{wu2023dipmark}, which significantly reduces their detection efficiency compared to other methods.

\subsection{Robustness}
 We compare the robustness of \methodname\ ($l=20$) with the SOTA unbiased watermark approaches ITS-edit, EXP-edit, $\gamma$-reweight, DiPmark, and STA-1. In this context, we use the text generation task with 1,000 generated sequences on LLaMA-2. For robustness evaluation, we manipulate $\epsilon\in\{0.05, 0.1, 0.2\}$ portion of the text tokens through token replacement attack~\citep{kirchenbauer2023reliability}, GPT rephrasing attack, GPT back translation attack, and DIPPER~\citep{krishna2023paraphrasing}.

 In Table~\ref{tab:robustness},~\ref{tab:gpt_rephrase_detection},~\ref{tab:gpt_back_translation_detection} and~\ref{tab:dipper_detection}, we present the TPR at a specific FPR and the median p-value for various watermarks across different attack strengths, denoted by $\epsilon$. \methodname\ consistently demonstrates superior robustness, outperforming all baseline methods in effectively detecting watermarked sentences.

\subsection{Detectability-robustness trade-off}
In Section~\ref{sec:trade-off}, we discuss the detectability-robustness trade-off of \methodname\ w.r.t. $l$. In this experiment, we empirically verify this trade-off by comparing the TPR@FPR=0.1\% and the median p-value with the number of the distribution channels $l\in$\{2, 3, 4, 5, 10, 20, 50, 100, 200, 500, 1000, 2000, 4000, 8000, 16000, 32000\}. We use Llama2 model with C4 subset on the text generation task. For robustness evaluation, we modify $\epsilon\in\{0.05, 0.1, 0.2\}$ portion of the text tokens through token replacement attack~\cite{kirchenbauer2023reliability,wu2023dipmark,chen2024watermark}.

In Figures~\ref{fig:trade-off1} and \ref{fig:trade-off2}, we report on the detectability of \methodname\ using two metrics: TPR@FPR=0.1\% and median p-value. The robustness of \methodname\ initially increases with $l$ before subsequently decreasing, aligning with the analysis presented in Section~\ref{sec:trade-off}.
\section{Conclusion}
In summary, we present \methodname, a novel family of unbiased watermarks that significantly enhance detectability and robustness in large language models without distorting text output. Our experimental results demonstrate a more than 10\% improvement in detectability over existing state-of-the-art unbiased watermarking approaches, validated across various models and datasets.  \methodname\ represents a significant advancement in the practical application of watermarking technology in LMs.

\section*{Limitations}
 There is an inherent trade-off between the robustness and detectability of our proposed \methodname. Increasing the number of distribution channels $l$ can potentially enhance detectability but reduce robustness against attacks where an adversary may modify output tokens to disrupt the watermark detection.

\section*{Acknowledgments}
This work was partially supported by NSF IIS 2347592, 2348169, DBI 2405416, CCF 2348306, CNS 2347617.

\bibliography{acl_latex}
\clearpage
\appendix

\section{Algorithms}

\begin{algorithm}[h]
\caption{Generation framework.}\label{alg:generation_framework}
\begin{algorithmic}[1]
\State \textbf{Input:} pretrained LM $P_M$, prompt $\bm{x}_{-m:0}$, generate length $T\in\mathbb{N}$.
\For{$t=1,\dots,T$}
        \State Get the probability distribution of $t$-th token $P_M(\cdot\mid\bm{x}_{-m:t-1})$.
        \State Generate the probability channels $\{P_1,...,P_l\}$
        \State Pseudorandomly select a probability channel $P_i$ based on the watermark key.
    \State Sample the next token $x_{i}$ from $P_i$.
\EndFor
\State \textbf{return} $\bm{x}_{1:T}$.
\end{algorithmic}
\end{algorithm}

\begin{algorithm}[h]
\caption{Detection framework.}\label{alg:detection_framework}
\begin{algorithmic}[1]
\State \textbf{Input:} pretrained LM $P_M$, generate length $T\in\mathbb{N}$, generated tokens $\bm{x}_{1:T}$.
\For{$t=1,\dots,T$}
        \State Recover the index of the selected distribution channel $i_t$ based on the watermark key.
    \State Check whether token $x_i$ is generated from the $i_t$-th distribution channel.
\EndFor
\end{algorithmic}
\end{algorithm}

\section{Robustness-detectability trade-off}\label{sec:full version}

An adversary may attempt to alter the output token to disrupt the watermark detection. In \methodname\ detection, if a token $x_t$ is modified to $x_t'$ and $x_t' \notin V_{i_t}$, the watermark signal is effectively removed. Consequently, the probability that a watermark is removed due to such an alteration is given by $\frac{|V_{i_t}|}{|V|}$.
Given that $|V_{i_t}| = \frac{|V|}{l}$, the probability that a watermark is removed simplifies to $\frac{1}{l}$
Therefore, increasing $l$ decreases the robustness of the watermark, as it increases the likelihood that an adversary can successfully remove the watermark by modifying the token.

On the other hand, moderately increasing \( l \) can enhance the detectability of the watermark.  Recall that $\Pr(\Phi(\x_{1:T}) \geq z) = \sum_{i = \lceil z \rceil}^{T} \binom{T}{i} \left(\frac{1}{l}\right)^i \left(\frac{l-1}{l}\right)^{T-i}.$
The derivative with respect to \( l \) of each term in the sum is given by:
$\frac{d}{dl} \left(\left(\frac{1}{l}\right)^i \left(\frac{l-1}{l}\right)^{T-i}\right) = \left(\frac{1}{l}\right)^{i+2} \left(\frac{l-1}{l}\right)^{T-i} \left(-il + (T-i)\frac{l}{l-1}\right),$
indicating that when \( l \geq 2 \) and \( i > \frac{T}{2} \), \( \left(\frac{1}{l}\right)^i \left(\frac{l-1}{l}\right)^{T-i} \) decreases with increasing \( l \). Typically, the score \( \Phi(\x_{1:T}) \) is greater than \( \frac{T}{2} \), so increasing \( l \) may reduce the detection p-value. This reduction makes the statistical distinction between watermarked and unwatermarked text more significant, thereby improving the detectability of the watermark. Thus, we identify a fundamental trade-off: increasing the number of distribution channels $l$ enhances the detectability of the watermark, yet it simultaneously reduces its robustness.

However, blindly increasing $l$ may also lead to bad detectability. The detectability is not only related to the distribution of $\Phi(\x_{1:n})$ under the null hypothesis but also the scale of $\Phi(\x_{1:n})$. If $l$ is too large, the expected true negative rate
$E_{TN}^{\methodname} = \sum_{i=1}^l \max\{0, 1/l - p_{V_i}\}
$ might be poor, since $p_{V_i}$ are more likely to be unevenly distributed. We empirically validate our analysis in Figure~\ref{fig:trade-off1} and ~\ref{fig:trade-off2}.

\section{Experimental settings}~\label{sec:settings}

\paragraph{Baselines} We evaluate the performance of our methods against various baselines, including two biased watermarking approaches, KGW~\citep{kirchenbauer2023watermark} and Unigram~\citep{zhao2023provable}, as well as five unbiased watermarking algorithms, ITS-edit~\citep{kuditipudi2023robust}, EXP-edit~\citep{kuditipudi2023robust}, $\gamma$-reweight~\citep{hu2023unbiased}, DiPmark~\citep{wu2023dipmark} and STA-1~\citep{mao2024watermark}.

\paragraph{Models and Datasets}

we ustilize Llama-2-7b-chat-hf~\citep{touvron2023llama}, Llama-3.2-3B-Instruct~\citep{dubey2024llama}, Mistral-7B-Instruct-v0.3~\citep{jiang2023mistral}, Phi-3.5-mini-instruct~\cite{abdin2024phi}for text generation tasks to evaluate the effectiveness of our proposed \methodname.

Following \citet{kirchenbauer2023watermark,hu2023unbiased}, we use a subset from the C4 dataset~\citep{raffel2020exploring} for text generation experiments. Additionally, we also include three MMW datasets~\citep{piet2023mark}, Dolly CW~\citep{DatabricksBlog2023DollyV2} and two tasks from WaterBench~\citep{tu2023waterbench}.

For unbiasedness validation, we adopt the settings from~\citet{hu2023unbiased,wu2023dipmark}, employing MBart~\citep{liu2020multilingual} for machine translation and BART~\cite{lewis2019bart} for text summarization.

In the machine translation experiments, we use the WMT16 ro-en dataset~\citep{bojar-EtAl:2016:WMT1}. For text summarization, while for text summarization, we utilize the CNN/DailyMail dataset~\citep{see-etal-2017-get}.

\paragraph{Watermarking parameters.} We evaluate the detectability of \methodname\ on the text generation task with different language models. We generate 1,000 examples for each tasks. We use the prefix 2-gram together with a secret key as the watermark keys. We select $\alpha \in\{ 0.3, 0.4\}$ for DiPmark, and $\delta \in \{0.5,1.0, 1.5, 2.0\}$ and $\gamma=0.5$ for KGW watermark \citep{kirchenbauer2023watermark}, $\delta \in \{0.5,1.0, 1.5, 2.0\}$ for Unigram~\citep{zhao2023provable}. For ITS-edit~\citep{kuditipudi2023robust}, EXP-edit~\citep{kuditipudi2023robust}, $\gamma$-reweight~\citep{hu2023unbiased} and STA-1~\citep{mao2024watermark}, we follow the settings in the original papers. For \methodname\ we set the number of distribution channels $l=20$.

\section{Additional Experiments}
In this section, we provide additional comparative analysis regarding the unbiased property and detectability of \methodname. We also include an ablation study on the number of distribution channels $l$ in \methodname.

\noindent\textbf{Unbiased Property.} In Tables~\ref{tab:unbiased_ts} and ~\ref{tab:unbiased_mt}, we conduct an additional evaluation of unbiasedness for both biased and unbiased watermarks. The results confirm that \methodname\ effectively preserves the language model's distribution, outperforming the biased watermark alternatives.

\begin{figure*}
    \centering
    \includegraphics[width=1\linewidth]{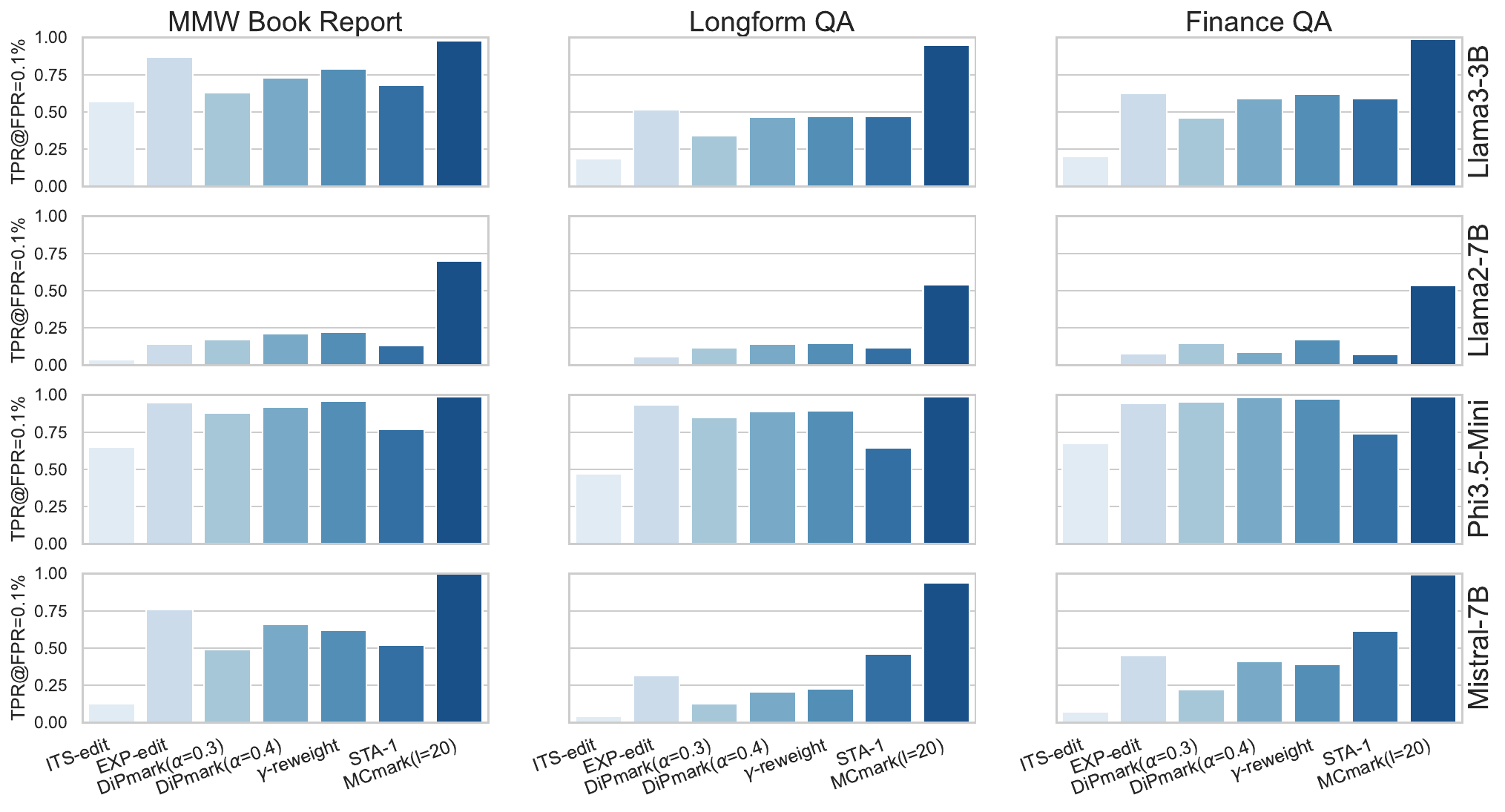}
    \caption{Comparative analysis of \methodname\ against SOTA unbiased watermarks across various language models and datasets on watermark detectability.}
    \label{fig:detectability comparison2}
\end{figure*}

\noindent\textbf{Detectability.} In Figure~\ref{fig:detectability comparison2}, we assess the detectability of \methodname\ on tasks such as MMW Book Report, Longform QA, and Finance QA. The results demonstrate that \methodname\ consistently exhibits superior detectability across all tested models and datasets.

\begin{figure*}[htbp]
  \centering
  \begin{subfigure}{0.48\textwidth}
    \centering
    \includegraphics[width=\textwidth]{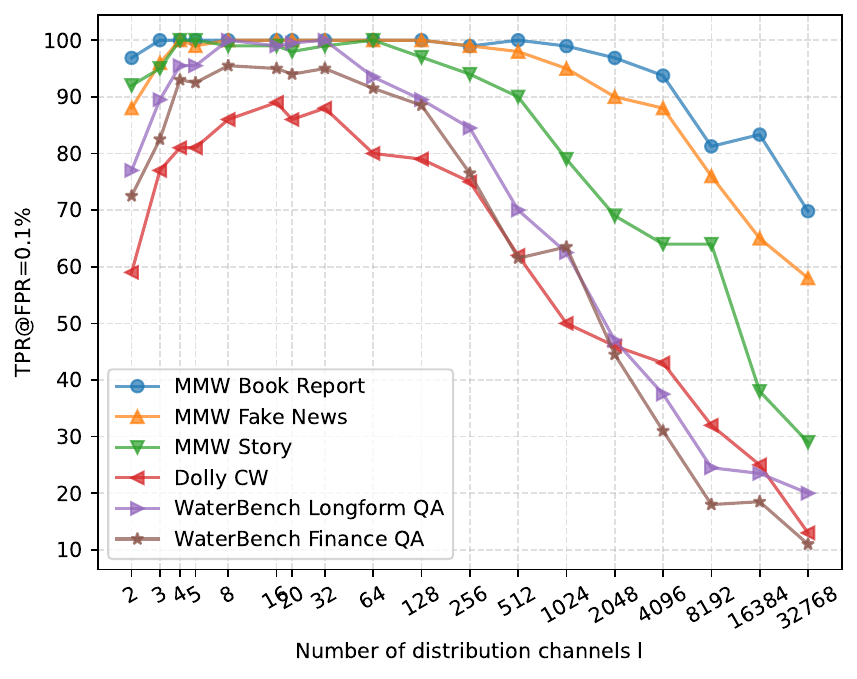}
  \end{subfigure}
  \quad
  \begin{subfigure}{0.48\textwidth}
    \centering
    \includegraphics[width=\textwidth]{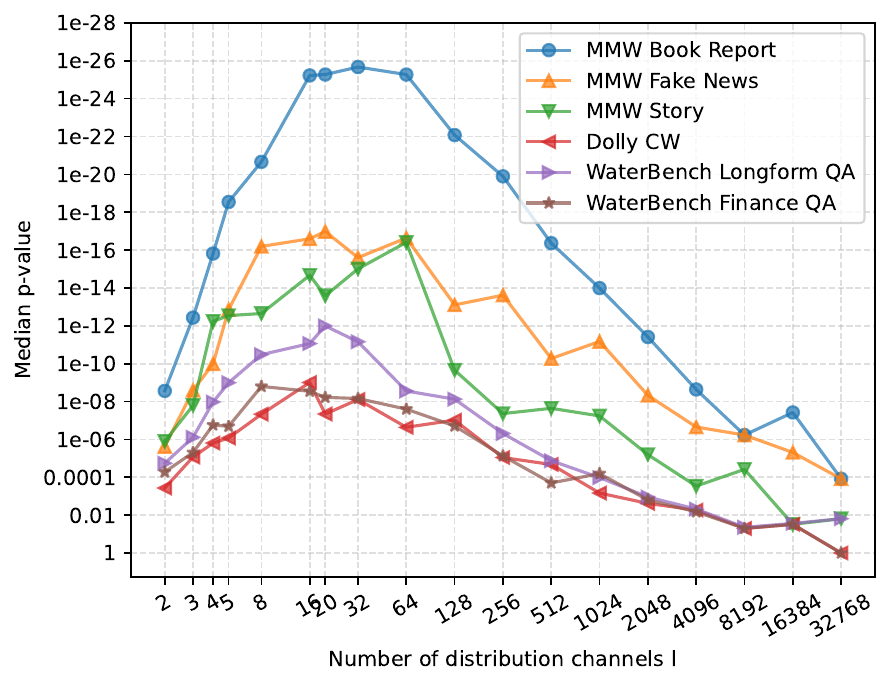}
  \end{subfigure}
  \caption{\textbf{Left}: Median p-value vs number of distribution channels $l$ in \methodname\ with Mistral-7B. \textbf{Right}: TPR@FPR=0.1\% vs number of distribution channels $l$ in \methodname\ with Mistral-7B.}
  \label{fig:Mistral-7B}
\end{figure*}

\begin{figure*}[htbp]
  \centering
  \begin{subfigure}{0.48\textwidth}
    \centering
    \includegraphics[width=\textwidth]{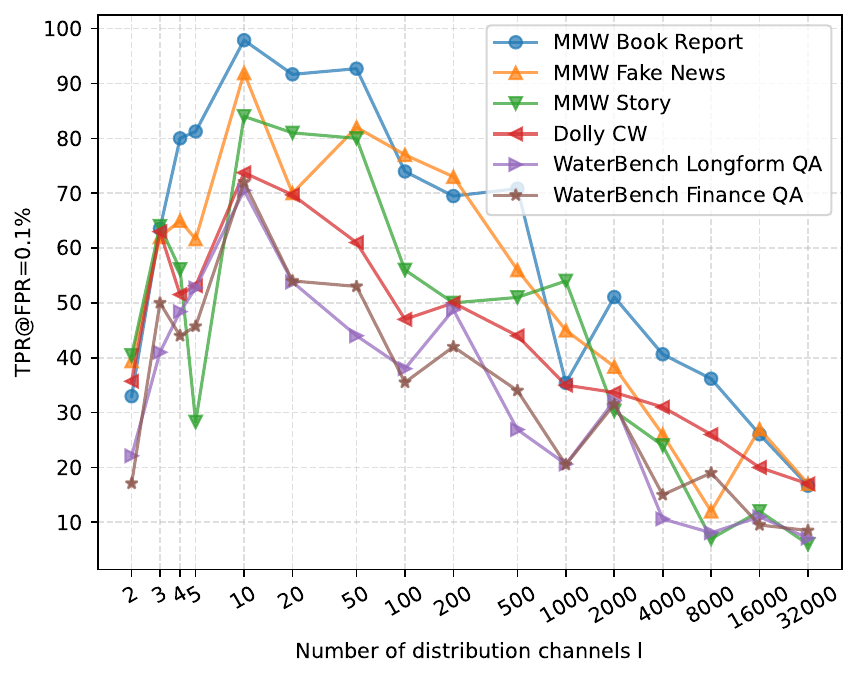}
  \end{subfigure}
  \quad
  \begin{subfigure}{0.48\textwidth}
    \centering
    \includegraphics[width=\textwidth]{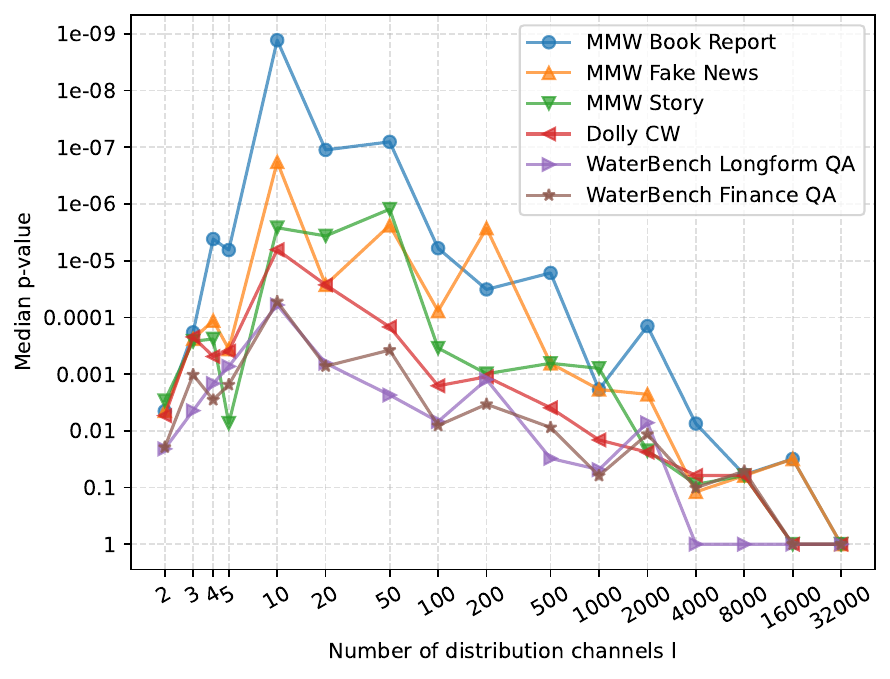}
  \end{subfigure}
  \caption{\textbf{Left}: Median p-value vs number of distribution channels $l$ in \methodname\ with Llama2-7B. \textbf{Right}: TPR@FPR=0.1\% vs number of distribution channels $l$ in \methodname\ with Llama2-7B.}
  \label{fig: Llama2-7B}
\end{figure*}

\begin{figure*}[htbp]
  \centering
  \begin{subfigure}{0.48\textwidth}
    \centering
    \includegraphics[width=\textwidth]{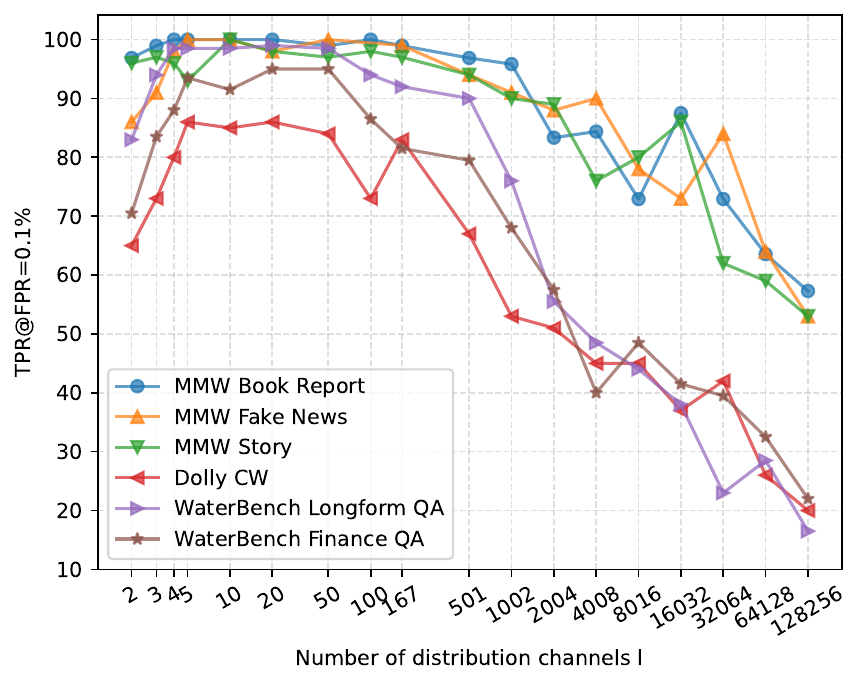}
  \end{subfigure}
  \quad
  \begin{subfigure}{0.48\textwidth}
    \centering
    \includegraphics[width=\textwidth]{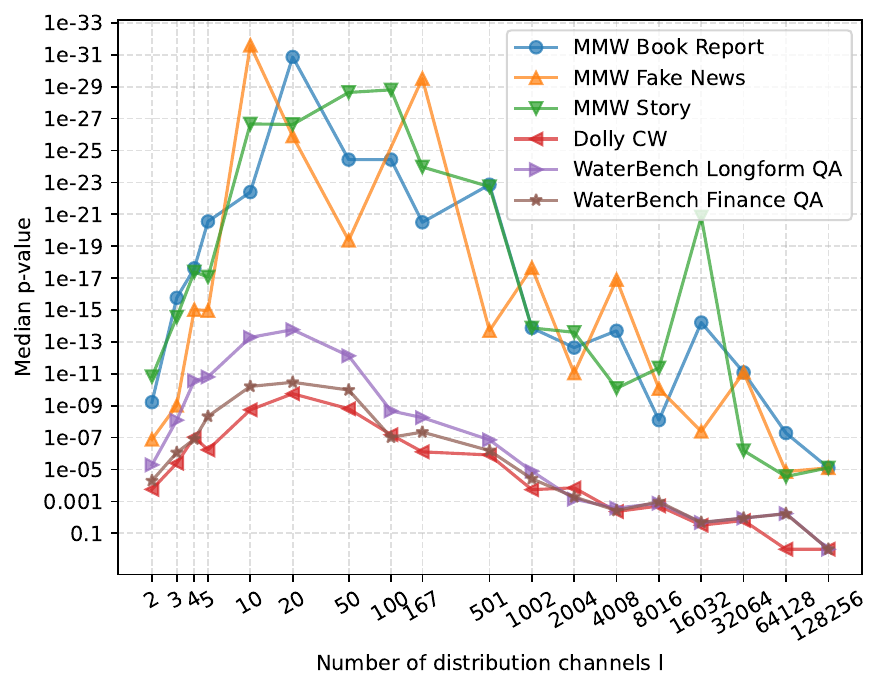}
  \end{subfigure}
  \caption{\textbf{Left}: Median p-value vs number of distribution channels $l$ in \methodname\ with Llama3-7B. \textbf{Right}: TPR@FPR=0.1\% vs number of distribution channels $l$ in \methodname\ with Llama3-7B.}
  \label{fig: Llama3-7B}
\end{figure*}

\begin{figure*}[htbp]
  \centering
  \begin{subfigure}{0.48\textwidth}
    \centering
    \includegraphics[width=\textwidth]{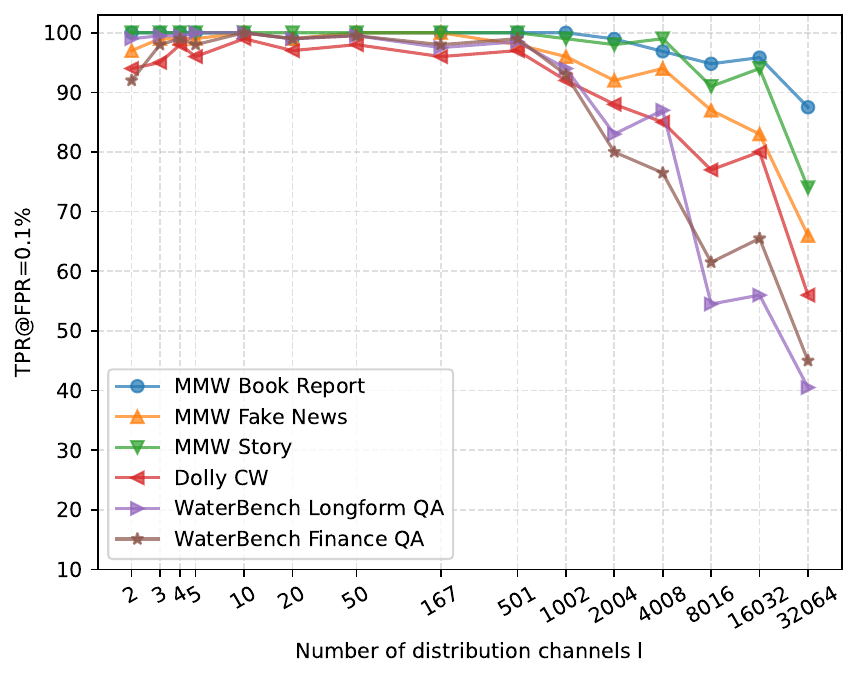}
    \caption{TPR@FPR=0.1\% vs $l$}
  \end{subfigure}
  \quad
  \begin{subfigure}{0.48\textwidth}
    \centering
    \includegraphics[width=\textwidth]{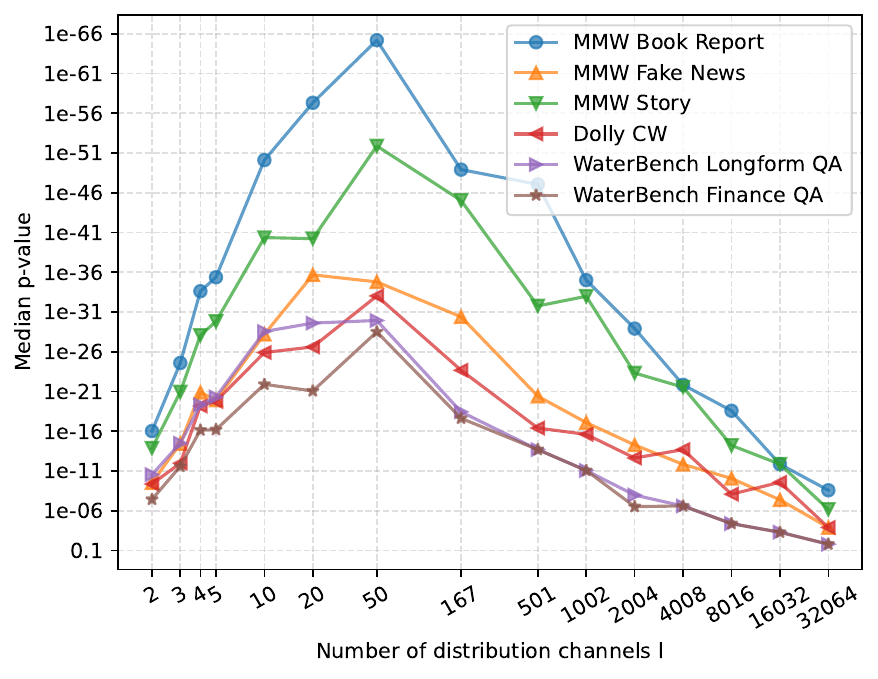}
    \caption{Median p-value vs $l$}
  \end{subfigure}
  \caption{\textbf{Left}: Median p-value vs number of distribution channels $l$ in \methodname\ with Phi3.5-Mini. \textbf{Right}: TPR@FPR=0.1\% vs number of distribution channels $l$ in \methodname\ with Phi3.5-Mini.}
  \label{fig: Phi3.5}
\end{figure*}

\noindent\textbf{Ablation Study with $l$.} In Figures~\ref{fig:Mistral-7B}, ~\ref{fig: Llama2-7B}, ~\ref{fig: Llama3-7B}, and ~\ref{fig: Phi3.5}, we present an analysis of the relationship between detectability and the number of distribution channels $l$ in \methodname. Our findings indicate that detectability initially increases and then decreases with respect to $l$, illustrating a critical trade-off in the parameter's configuration.

\begin{table}[htb]
\centering
\caption{Unbiasedness evaluation on text summarization tasks.}
\label{tab:unbiased_ts}
\resizebox{\columnwidth}{!}{
\begin{tabular}{l|ccc}
\toprule
 & BERT Score$\uparrow$ & PPL$\downarrow$ & Rouge-1$\uparrow$ \\ \midrule
Baseline & 0.3175 & 6.3932 & 0.3768 \\ \midrule
KGW($\delta$=0.5) & 0.3152 & 6.4894 & 0.3746 \\
KGW($\delta$=1.0) & 0.3125 & 6.8647 & 0.3742 \\
KGW($\delta$=1.5) & 0.3067 & 7.4633 & 0.3673 \\
KGW($\delta$=2.0) & 0.2996 & 8.4847 & 0.3605 \\ \midrule
Unigram($\delta$=0.5) & 0.3160 & 6.5302 & 0.3754 \\
Unigram($\delta$=1.0) & 0.3132 & 6.8145 & 0.3717 \\
Unigram($\delta$=1.5) & 0.3081 & 7.4693 & 0.3647 \\
Unigram($\delta$=2.0) & 0.2990 & 8.4182 & 0.3545 \\ \midrule
ITS-edit & 0.3147 & 6.5302 & 0.3758 \\
EXP-edit & 0.3209 & 5.9945 & 0.3775 \\
$\gamma$-reweight & 0.3164 & 6.4414 & 0.3765 \\
DiPmark($\alpha=0.4$) & 0.3178 & 6.4127 & 0.3773 \\
DiPmark($\alpha=0.3$) & 0.3169 & 6.3867 & 0.3765 \\
STA-1 & 0.3182 & 6.4118 & 0.3777 \\
\midrule
\methodname ($l$=20) &  0.3168& 6.3864 & 0.3763 \\ \bottomrule
\end{tabular}
}
\end{table}

\begin{table}[]
\centering
\caption{Unbiasedness evaluation on machine translation tasks.}
\label{tab:unbiased_mt}
\resizebox{\columnwidth}{!}{
\begin{tabular}{l|cc}
\toprule
 & BERT Score$\uparrow$ & BLEU$\uparrow$ \\ \midrule
Baseline & 0.5576 & 20.35 \\ \midrule
KGW($\delta$=0.5) & 0.5560 & 20.25 \\
KGW($\delta$=1.0) & 0.5555 & 20.02 \\
KGW($\delta$=1.5) & 0.5489 & 18.95 \\
KGW($\delta$=2.0) & 0.5420 & 18.28 \\ \midrule
Unigram($\delta$=0.5) & 0.5570 & 20.49 \\
Unigram($\delta$=1.0) & 0.5576 & 20.02 \\
Unigram($\delta$=1.5) & 0.5459 & 19.05 \\
Unigram($\delta$=2.0) & 0.5330 & 18.51 \\ \midrule
ITS-edit & 0.5700 & 21.29 \\
EXP-edit & 0.5600 & 20.00 \\
$\gamma$-reweight & 0.5548 & 20.12 \\
DiPmark($\alpha=0.4$) & 0.5614 & 20.65 \\
DiPmark($\alpha=0.3$) & 0.5563 & 20.48 \\
STA-1 & 0.5532 & 19.83 \\
\midrule
\methodname ($l$=20) & 0.5588 & 20.16 \\ \bottomrule
\end{tabular}
}
\end{table}

\end{document}